Proceedings of the ASME 2008 International Design Engineering Technical Conferences & Computers and Information in Engineering Conference
IDETC/CIE 2008
August 3-6, 2008, New York, USA

**DETC2008-49519.**# KINEMATIC ANALYSIS OF THE VERTEBRA OF AN EEL LIKE ROBOT

**Damien Chablat**
*Institut de Recherche en Communications et Cybernétique de Nantes*
*1, rue de la Noë, 44321 Nantes, France*
Damien.Chablat@irccyn.ec-nantes.fr## ABSTRACT

The kinematic analysis of a spherical wrist with parallel architecture is the object of this article. This study is part of a larger French project, which aims to design and to build an eel like robot to imitate the eel swimming. To implement direct and inverse kinematics on the control law of the prototype, we need to evaluate the workspace without any collisions between the different bodies. The tilt and torsion parameters are used to represent the workspace.## INTRODUCTION

Parallel kinematic architectures are commonly claimed to offer several advantages over their serial counterparts, like high structural rigidity, high dynamic capacities and high accuracy [1]. Thus, they are interesting for applications where these properties are needed, such as flight simulators [2] and high-speed machines. Recently, new applications have used such mechanisms to build humanoid robots [3] or snake robots [4].

Over millions of years, fish have evolved swimming capacity far superior in many ways to what has been by nautical science and technology. They use their streamlined bodies to exploit fluid-mechanical principles. This way, they can achieve extraordinary propulsion efficiencies, acceleration and maneuverability not feasible by the best naval architects [5].

In [6], we have introduced a new architecture of spherical wrist able to reproduce the vertebra of an eel. The purpose of this article is to show the kinematic equations and to study its workspace taking into account mechanical constraints to avoid internal collisions. The Euler angles are classically used to compute the workspace of spherical wrists but they do not permit to visualize the symmetrical properties. Recently in [7], the Tilt-and-Torsion was introduced to represent the workspace of the agile eye. In this paper, we will present the kinematic equations of the spherical wrist and an algorithm to compute the workspace and the joint space.

## PRELIMINARIES

### Biomimetic robotics

The aim of our project is to imitate the eel swimming and its biological systems and to conceive new technologies drawn from the lesson of their study [8].

Many researches have been made in the underwater field in America and in Japan [9]. In this context, two modes of locomotion mainly attract the attention of researchers, (i) the carangid swimming (family Carangidae as the one of a jack, a horse mackerel or a pompano [10]) based on oscillations of the body and (ii) the anguilliform swimming (of snake type, eel, lamprey, etc.) based on undulations of the body. An anguilliform swimmer propels itself forward by propagating waves of curvature backward along its body [5].

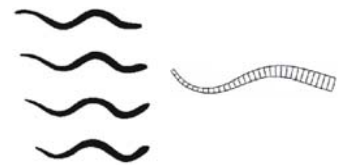

Figure 1: Change in body shape in swimming and a subdivision of its body

1Copyright © #### by ASME

To carry out anguilliform swimming, the body of the eel is made of a succession of vertebrae whose undulation produces motion, as depicted in Fig. 1. In nature, there is only one degree of freedom between each vertebra because the motion control of the vertebrae is coupled with the motion of the dorsal and ventral fin. These two fins being not easily reproducible, we will give to each vertebra more mobility to account problems of rolling, for example. The assembly of these vertebrae, coupled to a head having two fins, must allow the reproduction of the eel swimming.

From the observation of European eel, Anguilla anguilla, we have data concerning its kinematic swimming such as wave speed, cycle frequency, amplitude or local bending [11]. The yaw is given for forward and backward swimming on total body length, as depicted in Fig. 2. The other angles are obtained using Navier-Stokes equations on characteristic trajectories [12]. For our prototype, we took as constraints of design, ±30 degrees in yaw for forward swimming, ±15 degrees in pitching for diving and ±4 degrees in rolling to compensate for torsion in diving.

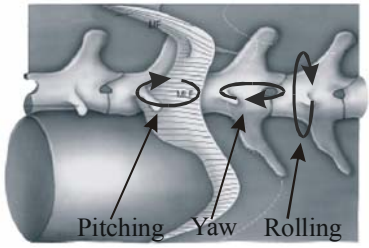

Figure 2: Rolling, pitching and yaw angles of vertebrae

### Orientation representation

The tilt-and-torsion (T&T) angles are defined in [7] as a combination of a tilt and a torsion. These angles are defined in two stages. In the first one, illustrated in Fig. 3(a), the body frame is tilted about a horizontal axis, $a$, at an angle $\theta$, referred to as the tilt. The axis $a$ is defined by an angle $\varphi$, called the azimuth, which is the angle between the projection of the body $z'$ axis onto the fixed $xy$ plane and the fixed $x$-axis. In the second stage, illustrated in Fig. 3(b), the body frame is rotated about the body $z'$ axis at an angle $\psi$, called the torsion.

$$R = R_z(\varphi) R_y(\theta) R_z(\psi - \varphi)$$

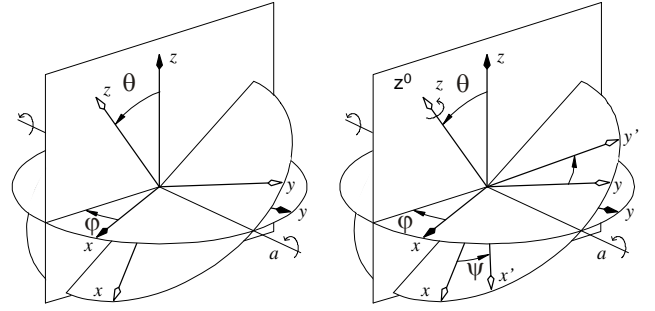

Figure 3: The successive rotations of the T&T angles: (a) tilt, (b) torsion.

For a given torsion angle, the workspace can be represented in a polar coordinate system as is shown in Fig. 4.

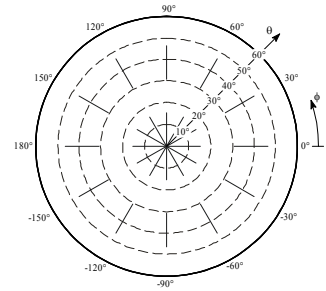

Figure 4: Projected orientation workspace in a polar coordinate system ($\phi$, $\theta$) for a given torsion $\psi$

## KINEMATIC STUDY OF THE SELECTED ARCHITECTURE

### Description

The architecture chosen in [6] is a non-overconstrained asymmetrical architecture that is reported in [16] as an (3, 6, 6) architecture. The base and the mobile platform are connected by three kinematic chains, as depicted in Fig. 5.

This architecture results from the research around the Lie Group of Euclidian displacements [15]. There are (i) two kinematic chains, noted legs ① and ②, to produce a general rigid body displacement from the subgroup {D} (6 DOF) and (ii) a kinematic chain, noted leg ③, from the spherical subgroup {S} and made by three coaxial revolute joints (3 DOF). There is only one actuated joint on each leg $(\theta_1, \theta_2, \theta_3)$.

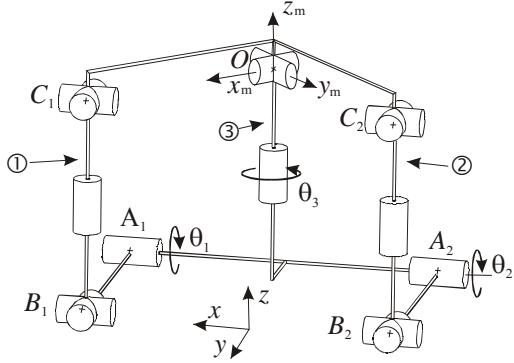

**Figure 5: Structure of the studied spherical wrist**

If the realization of leg ③ is easy (three coaxial revolute joints), it is difficult to enumerate all the legs with 6-DOF. The most current generator of {D} is of the UPS type (Gough-Stewart's platform, with P prismatic actuated joint, U for universal joint and S for spherical joint), which has the disadvantage of using a prismatic actuator that is not fixed on the basis.

In literature, an equivalent mechanism exists but the generator of {D} is of PUS type (with P prismatic actuated joint). For legs ① and ②, the prismatic actuated joints are in parallel to the vertebral column which is harmful for the compactness of the mechanism. The orientation can be changed but the efficiency decreases considerably. For leg ③, the first revolute joint (located on the base) is actuated.

Thus, we have changed the type of legs ① and ②, by a RUS type (with R revolute actuated joint) as depicted in Fig. 5.

In our project, we have built an eel robot with 10 vertebrae and an overall length of 1500 mm, as shown in Fig. 6 (with the head and the tail included). Each vertebra will have an elliptic section of 150 mm and 100 mm focal distance respectively and will be a 100 mm thick. The workspace analysis is needed to avoid collision and to check the joint limits.

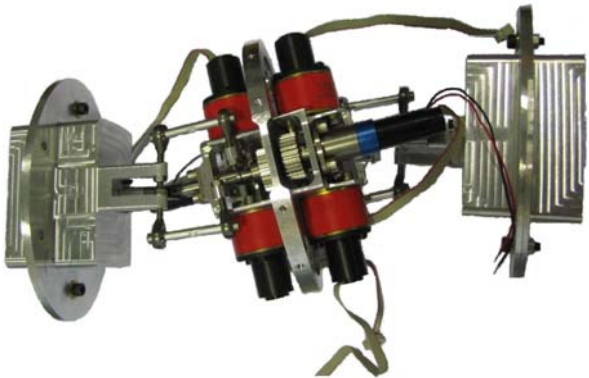

**Figure 6: Two vertebrae of the prototype under construction at IRCCyN**

For the prototype, we have replaced the joints located in $C_1$, $C_2$, $B_1$ and $B_2$ by spherical joints. The motor associated with the joint $\theta_3$ is not coaxial but parallel and two gears transmit the motion.

### Kinematics

A fixed reference frame, noted $\Re_{fixed}(O, x, y, z)$ is located on the base and is oriented in such a way that (i) plane $Oxz$ is defined by points $C_1$, $C_2$ and $O$, (ii) the z-axis is vertical, (iii) x-axis is directed from $A_2$ to $A_1$. The coordinates of points $A_1$ and $A_2$ in $\Re_{fixed}$ are written as

$$[OA_1]_{\Re_{fixed}} = \begin{bmatrix} a & b & c \end{bmatrix}^T \text{ and } [OA_2]_{\Re_{fixed}} = \begin{bmatrix} -a & b & c \end{bmatrix}^T \quad (1)$$

The lengths $a$, $b$, and $c$ will be chosen by the study of the Jacobian matrix in the next subsection.

The mobile platform will be rotating around point $O$ that is the origin of the mobile frame, noted $\Re_{mobile}$. The orientation of $\Re_{mobile}(O, x_m, y_m, z_m)$ is defined so that (i) plane $Ox_m y_m$ is the plane defined by points $C_1$, $C_2$ and $O$, (ii) $x_m$-axis is directed from $O$ to $C_1$ and (iii) $y_m$-axis is directed from $O$ to $C_2$.

Let $\theta$ be the vector of joint coordinates associated with the actuated revolute joints. The orientation of the mobile platform with respect to fixed frame $\Re_{base}$ is defined by the "Rolling Pitching Yaw" parameters (RPY) where the first parameter is the orientation angle $\theta_3$ of the first revolute joint of leg ③).

$$\theta = \begin{bmatrix} \theta_1 & \theta_2 & \theta_3 \end{bmatrix}^T$$

$$^{fixed}\mathbf{R}_{mobile} = \mathbf{R}(z, \theta_3) \mathbf{R}(y', \phi) \mathbf{R}(x'', \psi)$$

The angles $\theta_3$, $\phi$ and $\psi$ are associated with the following cascaded rotations

(i) a rotation of angle $\theta_3$ around z-axis,

(ii) a rotation of angle $\phi$ around the y'-axis (obtained from the previous rotation and whose axis is the axis of the second revolute joint of leg ③),

(iii) a rotation of $\psi$ around the x''-axis (obtained from the second rotation and whose axis is the axis of the third revolute joint of leg ③).

### Jacobian matrices

To characterize the singular configurations, we will use an invariant form, which allows our results to be applicable to any architecture studied here. Thus, there is no problem of singularity of transformation in the rotation matrix between $\Re_{fixed}$ and $\Re_{mobile}$.

 

We write the Chasles's relation on $(\mathbf{c}_i - \mathbf{b}_i)$ to have

$$(\mathbf{c}_i - \mathbf{b}_i) = (\mathbf{c}_i - \mathbf{o}_i) + (\mathbf{o}_i - \mathbf{a}_i) - (\mathbf{b}_i - \mathbf{a}_i) \quad (2)$$

In this equation, all the vectors are expressed in $\Re_{fixed}$. To simplify calculations, we set

$$\mathbf{r}_i = (\mathbf{c}_i - \mathbf{b}_i), \mathbf{p}_i = (\mathbf{c}_i - \mathbf{o}_i), \mathbf{b}_i = (\mathbf{o}_i - \mathbf{a}_i) \text{ and } \mathbf{l}_i = (\mathbf{b}_i - \mathbf{a}_i)$$

By differentiating Eq. (2) with respect to time, we obtain,

$$\dot{\mathbf{r}}_i = \dot{\mathbf{p}}_i - \dot{\mathbf{l}}_i \quad (3)$$

with

$$[\mathbf{p}_i]_{\Re_{fixed}} = {}^{fixed}\mathbf{R}_{mobile} [\mathbf{p}_i]_{\Re_{mobile}} \quad (4)$$

Differentiating with respect to time, we find

$$[\dot{\mathbf{p}}_i]_{\Re_{fixed}} = \dot{\mathbf{Q}} {}^{fixed}\mathbf{R}_{mobile} [\mathbf{p}_i]_{\Re_{mobile}} \quad (5)$$

since vector $[\mathbf{p}_i]_{\Re_{mobile}}$ is a constant vector when expressed in frame $\Re_{mobile}$. Moreover, the time derivation of the rotation matrix can be written as

$$\dot{\mathbf{Q}} = \mathbf{\Omega}\mathbf{Q} \quad (6)$$

where $\mathbf{\Omega}$ is the angular velocity tensor. Finally, from Eqs. (2) and (6), we get

$$\dot{\mathbf{p}}_i = \mathbf{\Omega}\mathbf{p}_i = \boldsymbol{\omega} \times \mathbf{p}_i$$

where $\times$ denotes the cross product of the two vectors and $\boldsymbol{\omega}$ is the angular velocity vector. We note $\mathbf{i}_1$ and $\mathbf{i}_2$, the unit vectors passing through the axis of the first revolute joint of legs ① and ②, respectively. Moreover, we can write vector $\dot{\mathbf{l}}_i$ as function of angular velocities $\dot{\theta}_1$ and $\dot{\theta}_2$

$$\dot{\mathbf{l}}_1 = \mathbf{l}_1 \times (\dot{\theta}_1 \mathbf{i}_1) \text{ and } \dot{\mathbf{l}}_2 = \mathbf{l}_2 \times (\dot{\theta}_2 \mathbf{i}_2)$$

Thus, Eq. (3) can be written in the form

$$\dot{\mathbf{r}}_i = \boldsymbol{\omega} \times \mathbf{p}_i - \mathbf{l}_i \times (\dot{\theta}_i \mathbf{i}_i)$$

We multiply the preceding equation by $\mathbf{r}_i^T$ because $\mathbf{r}_i^T \dot{\mathbf{r}}_i = 0$. Thus, we have

$$\mathbf{r}_i^T .(\boldsymbol{\omega} \times \mathbf{p}_i) = \mathbf{r}_i^T .(\mathbf{l}_i \times (\dot{\theta}_i \mathbf{i}_i))$$

Or

$$(\mathbf{p}_i \times \mathbf{r}_i)^T .\boldsymbol{\omega} = (\mathbf{l}_i \times \mathbf{r}_i)^T .(\dot{\theta}_i \mathbf{i}_i)$$

These two equations can be cast in vector form

$$\mathbf{A}\boldsymbol{\omega} + \mathbf{B}\dot{\mathbf{q}} = 0 \quad (7)$$

with

$$\mathbf{A} = \begin{bmatrix} (\mathbf{p}_1 \times \mathbf{r}_1)^T \\ (\mathbf{p}_2 \times \mathbf{r}_2)^T \\ 0 & 0 & 1 \end{bmatrix} \quad (8)$$

$$\mathbf{B} = \begin{bmatrix} (\mathbf{l}_1 \times \mathbf{r}_1)^T .\mathbf{i}_1 & 0 & 0 \\ 0 & (\mathbf{l}_2 \times \mathbf{r}_2)^T .\mathbf{i}_2 & 0 \\ 0 & 0 & 1 \end{bmatrix} \quad (9)$$

and $\dot{\mathbf{q}} = \begin{bmatrix} \dot{\theta}_1 & \dot{\theta}_2 & \dot{\theta}_3 \end{bmatrix}^T$

Then, when $\mathbf{B}$ is not singular, the inverse Jacobian matrix is written,

$$\mathbf{J}^{-1} = \begin{bmatrix} \dfrac{1}{(\mathbf{l}_1 \times \mathbf{r}_1)^T .\mathbf{i}_1} (\mathbf{p}_1 \times \mathbf{r}_1)^T \\ \dfrac{1}{(\mathbf{l}_2 \times \mathbf{r}_2)^T .\mathbf{i}_2} (\mathbf{p}_2 \times \mathbf{r}_2)^T \\ 0 & 0 & 1 \end{bmatrix}$$

**Singular configurations**

The parallel singularities occur when the determinant of the matrix $\mathbf{A}$ vanishes, *i.e.* when $\det(\mathbf{A}) = 0$. In such configurations, it is possible to move locally the mobile platform whereas the actuated joints are locked. These singularities are particularly undesirable because the structure cannot resist any force or torque.

From Eq. (7), we have

$$(\mathbf{p}_1 \times \mathbf{r}_1) \square (\mathbf{p}_2 \times \mathbf{r}_2) \text{ or } (\mathbf{p}_1 \times \mathbf{r}_1) = 0 \text{ or } (\mathbf{p}_2 \times \mathbf{r}_2) = 0$$

           

It is equivalent to have $B_1$, $B_2$, $C_1$, $C_2$ and $O$ coplanar or to have ($B_1$, $C_1$, $O$) or ($B_2$, $C_2$, $O$) aligned, as depicted in Fig. 7.

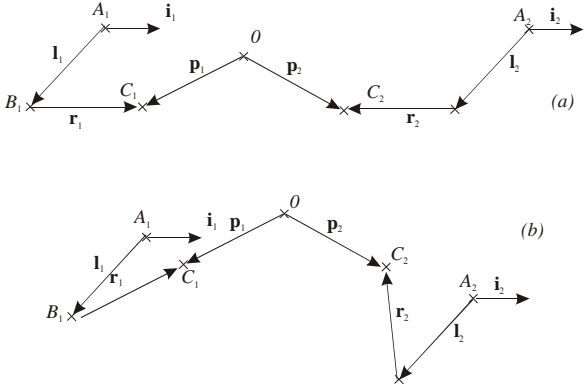

**Figure 7: Parallel singularity when (a) $B_1$, $B_2$, $C_1$, $C_2$ and $O$ are coplanar and (b) $B_1$, $C_1$ and $O$ are aligned**

Serial singularities occur when the determinant of the matrix **B** vanishes, *i.e.* when $\det(\mathbf{B}) = 0$. At a serial singularity, an orientation exists along which any angular velocity cannot be produced.
From Eq. (8), we have

$$(\mathbf{l}_1 \times \mathbf{r}_1)^T \mathbf{i}_1 = 0 \text{ or } (\mathbf{l}_2 \times \mathbf{r}_2)^T \mathbf{i}_2 = 0 \text{ or}$$

$$(\mathbf{i}_1 \times \mathbf{r}_1)^T \mathbf{l}_1 = 0 \text{ or } (\mathbf{i}_2 \times \mathbf{r}_2)^T \mathbf{l}_2 = 0$$

It is equivalent to have (i) $\mathbf{l}_1$ and $\mathbf{r}_1$ aligned, or (ii) $\mathbf{l}_2$ and $\mathbf{r}_2$ aligned, or (iii) $\mathbf{r}_1$ and $\mathbf{i}_1$ aligned, or (iv) $\mathbf{r}_2$ and $\mathbf{i}_2$ aligned, as depicted in Fig.8.

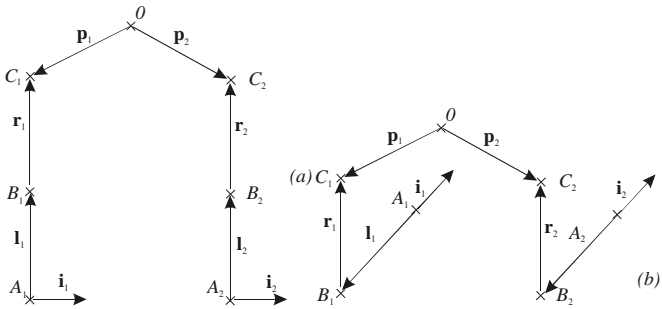

**Figure 8: Serial singularity when
(a) $\mathbf{l}_i$ and $\mathbf{r}_i$ are aligned and (b) $\mathbf{l}_i$ and $\mathbf{i}_i$ are aligned**

## Condition number and isotropic configurations

The Jacobian matrix is said to be isotropic when its condition number attains its minimum value of one [20]. The condition number of the Jacobian matrix is an interesting performance index, which characterizes the distortion of a unit ball under the transformation represented by the Jacobian matrix. The matrix **A** is isotropic when

$$\mathbf{p}_1 \perp \mathbf{r}_1 \text{ and } \mathbf{p}_2 \perp \mathbf{r}_2 \text{ and}$$

$$(\mathbf{p}_1 \times \mathbf{r}_1) \perp (\mathbf{p}_2 \times \mathbf{r}_2) \text{ and } \|\mathbf{r}_1\| = \|\mathbf{p}_1\| = \|\mathbf{r}_2\| = \|\mathbf{p}_2\| = 1$$

The matrix **B** is isotropic and equal to the identity matrix when

$$\mathbf{l}_1 \perp \mathbf{r}_1 \text{ and } \mathbf{l}_2 \perp \mathbf{r}_2 \text{ and } \mathbf{l}_1 \perp \mathbf{i}_1 \text{ and } \mathbf{l}_2 \perp \mathbf{i}_2 \text{ and } \|\mathbf{r}_i\| = \|\mathbf{l}_i\| = 1.$$

From these conditions, in [6], we have isolated three cases.
• The first solution is the mechanism depicted in Fig. 5 that we could call "parallel axes". Equation 1 gives the location of points $A_1$ and $A_2$ in $\Re_{base}$ for a unit mechanism,

$$a = \frac{\sqrt{2}}{2}, \quad b = \frac{\sqrt{2}-2}{2}, \quad c = -1$$

If this solution admits an isotropic configuration, the behavior in forward swimming leads to use legs ① and ② simultaneously. When we apply as input velocity $\dot{\mathbf{\theta}} = \begin{bmatrix} 1 & 1 & 0 \end{bmatrix}^T$, the angular velocity obtained is $\mathbf{\omega} = [\sqrt{2} \quad 0 \quad 0]^T$. This means that we amplify the rotational motion just after having used a reduction gear on the rotary motor to increase the available torque. Thus, the length of the motors is constrained by the shape of the cross section of the eel, as depicted in Fig. 10 (a).
• The second solution, called "orthogonal axes", is to place $\mathbf{i}_1$ and $\mathbf{i}_2$ orthogonally as depicted in Fig. 9. The location of points $A_1$ and $A_2$ in $\Re_{base}$ coincides with point $O$.

 

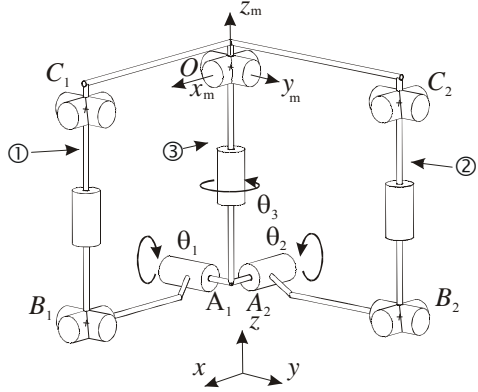

**Figure 9: Spherical wrist with orthogonal actuators**

In this case, the direct and inverse kinematic models are simpler but it is more difficult to place the motors of legs ① and ②, as shown in Fig. 10 (b). Moreover, there also exists an angular amplification factor in the forward swimming.

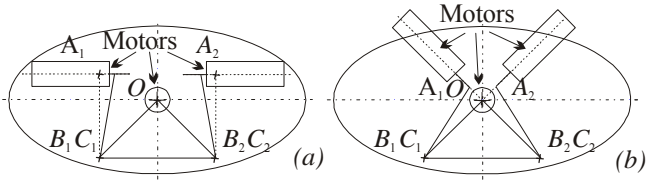

**Figure 10: Placement of the motors and the legs for (a) the "parallel axes" and (b) the "orthogonal axes"**

- The last solution has parallel actuators and their axes intersect the *z*-axis, as depicted in Fig. 11. When the eel robot is swimming, the angular velocity of the actuated joints of legs ① and ② is equal to yaw velocity.

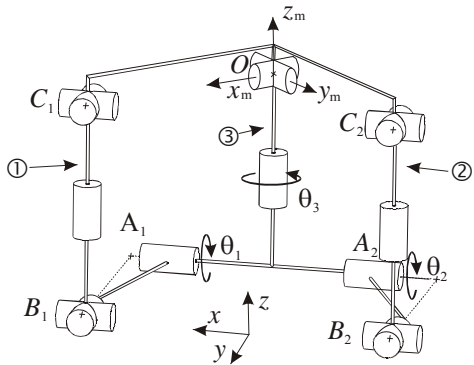

**Figure 11: Spherical wrist with parallel actuators**

This means that for the forward or backward swimming, the kinematic models are simple and the torque needed for the motion is distributed. However, only **A** can be isotropic because we have

$$(\mathbf{l}_i \times \mathbf{r}_i)^T \cdot \mathbf{i}_i = \sqrt{2}/2 \quad \text{for i=1,2}$$

Equation 1 gives the location of points $A_1$ and $A_2$ in $\Re_{base}$ for a unit mechanism,

$$a = \sqrt{2}/2, \quad b = 0, \quad c = -1$$

Concerning the integration into the cross-section of the eel, the placement is less constrained, as shown in Fig. 12.

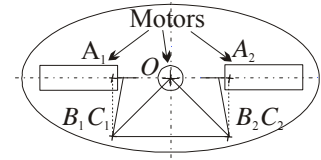

**Figure 12: Placement of the motors and the legs for spherical wrist with parallel actuators**

## Direct and inverse kinematic models

The direct kinematic model can be written when we know the position of $B_i$ and $C_i$. Thus, we have,

$$B_1 = \left[ \frac{\sqrt{2}}{2} \quad \frac{\sqrt{2}C_1}{2} \quad -1 + \frac{\sqrt{2}S_1}{2} \right]^T \quad B_2 = \left[ -\frac{\sqrt{2}}{2} \quad \frac{\sqrt{2}C_2}{2} \quad -1 + \frac{\sqrt{2}S_2}{2} \right]^T$$

and in $\Re_{mobile}$,

$$C_1 = \begin{bmatrix} 1 & 0 & 0 \end{bmatrix}^T \quad C_2 = \begin{bmatrix} 0 & 1 & 0 \end{bmatrix}^T$$

or in $\Re_{fixed}$

$$C_1 = \begin{bmatrix} C_3 C_\phi \\ S_3 C_\phi \\ -S_\phi \end{bmatrix} \quad C_2 = \begin{bmatrix} C_3 S_\phi S_\psi - S_3 C_\psi \\ S_3 S_\phi S_\psi + C_3 C_\psi \\ C_\phi S_\psi \end{bmatrix}$$

with $C_i = \cos(\theta_i)$, $S_i = \sin(\theta_i)$ for i=1,2,3, $C_\phi = \cos(\phi)$, $S_\phi = \sin(\phi)$, $C_\psi = \cos(\psi)$ and $S_\psi = \sin(\psi)$.

We add the constraint that $\|B_i C_i\| = 1$

 

$$\left| C_3 C_\phi - \frac{\sqrt{2}}{2} \right|^2 + \left| S_3 C_\phi - \frac{\sqrt{2}}{2} C_1 \right|^2 + \left| S_\phi - 1 + \frac{\sqrt{2}}{2} S_1 \right|^2 = 1 \quad (10)$$

$$\left| C_3 S_\phi S_\psi - S_3 C_\psi + \frac{\sqrt{2}}{2} \right|^2 + \left| S_3 S_\phi S_\psi + C_3 C_\psi - \frac{\sqrt{2}}{2} C_2 \right|^2 + \left| \frac{\sqrt{2}}{2} S_2 - C_\phi S_\psi - 1 \right|^2 = 1 \quad (11)$$

To solve the direct kinematic, we know $\mathbf{\theta} = [\theta_1 \; \theta_2 \; \theta_3]^T$ and we use the following substitutions

$$\sin(\phi) = \frac{2Q}{1+Q^2} \quad \cos(\phi) = \frac{1-Q^2}{1+Q^2}$$

Thus, we can remark that Eq. 10 depends only on $\phi$ and is a quadratic equation of Q

$(\sqrt{2} S_1 Q - \sqrt{2} Q C_3 - \sqrt{2} Q C_1 S_3 - 2Q + 2 - \sqrt{2} C_1 S_3 - \sqrt{2} C_3 - \sqrt{2} S_1)$
$(Q - 1) = 0$

One solution is $Q = 1$, i.e. $\phi = \pi/2 + 2k\pi$ that does not depend on the actuated joints. Figure 13 depicts the four direct kinematic solutions for $\theta_1 = 0.1$, $\theta_2 = 0.2$, $\theta_3 = \pi/4$. Solutions (a) and (b) are found when $Q = 1$ and can be easily isolated. From solutions (c) and (d), only the second one is suitable, it can be isolated by the dot product of $\mathbf{r}_2$ by $\mathbf{p}_2$.

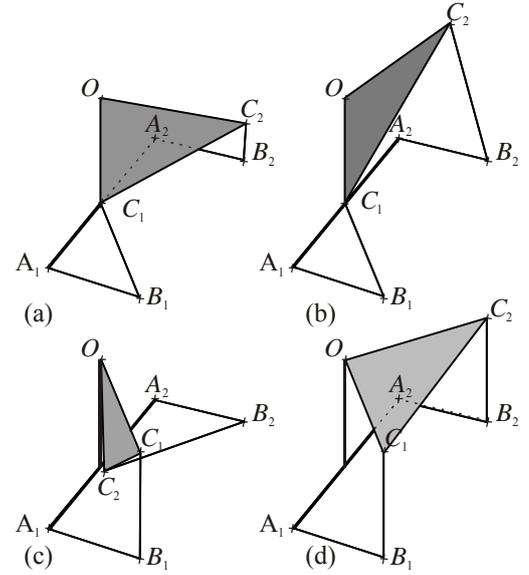

**Figure 13: The four direct kinematic solutions for**
$\theta_1 = 0.1, \; \theta_2 = 0.2, \; \theta_3 = \pi/4$

To solve the inverse kinematic, we use two substitutions, $R = \tan(\theta_1/2)$ and $S = \tan(\theta_2/2)$ that allow us to have two quadratic and independent equations as function of R and S respectively. Figure 14 shows the four inverse kinematic solutions for $\theta_3 = \pi/4$, $\phi = \pi/12$, $\psi = \pi/12$ that we can isolate by calculating $\mathbf{l}_1 . \mathbf{r}_1$ and $\mathbf{l}_2 . \mathbf{r}_2$ for legs ① and ②, respectively.

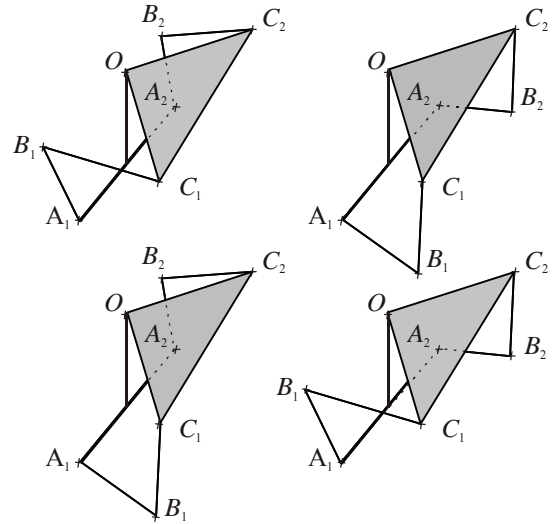

**Figure 14: The four inverse kinematic solutions for**
$\theta_3 = \pi/4, \; \phi = \pi/12, \; \psi = \pi/12$

    

To conclude, we have four solutions for the direct kinematic and four solutions for the inverse kinematic (two for legs ① and ②, respectively).

## Mechanical constraints

To explain the mechanical constraints, we use a simplified model depicted on Fig.15 but the numerical constraints come from the real prototype shown in Fig. 6.

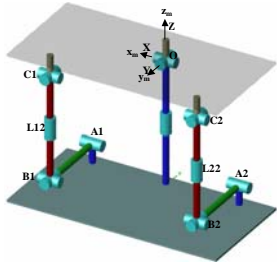

**Figure 15: Simplified model of a vertebra**

For a parallel manipulator, the most common constraints are:
- The joint constraints that take into account the variation of the length of the legs that contain slides, which is not our case.
- The constraints on the joint limits, which represent the constraints on the spherical joints.
- The intersection between segments, which will be the intersection of $(B_1C_1)$ and $(B_2C_2)$ or $(B_iC_i)$ and the leg ③ (the latter is impossible because of the mechanism architecture).
- The intersection segments with obstacles, which will be defined for our system by the intersection of rods with the low base.

We will study the three constraints that exist for the mechanism later on.

### The joint limits on universal joints

Our spherical joints are limited by a 'lima' so $C_iB_i$ will be required within the cones (Fig. 16):
- The cone center $B_i$ and a half angle 'Lima' and having its axis of symmetry perpendicular to the rod $A_iB_i$.
- The cone center $C_i$ and a half angle 'Lima' and having its axis of symmetry perpendicular to the mobile platform.

These limits have the most influence on the size of the workspace.

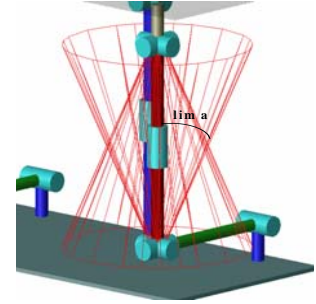

**Figure 16: Joint limits on universal joints**

### The intersection between segments

To avoid these intersections, we have limited the minimum distance between segments. The methodology for finding the minimum distance is as follows:
- We take a point $M_1$, which belongs to $B_1C_1$.
- Take $M_2$ projection of $M_1$ on $(B_2C_2)$, we measure the distances $\|M_2 B_2\|$ and $\|M_2 C_2\|$.
- If these two distances are less than the length of $(B_2C_2)$, we define the distance between the two segments as distance $M_1M_2$.
- We are changing the position of the $M_1$ throughout the segment $B_1C_1$ and the minimum distance between segments will be the minimum length $M_1M_2$.

This method requires a large computation time and there is little risk to meet this configuration when the other constraints are validated.

### The intersection between segments and the base

The only way to have internal collisions between a segment and a fixed part is to have an intersection between segments $(A_iB_i)$ and the base. To eliminate them, we limited the angle $\theta_i$ (i= 1, 2) as $\sin(\theta_i)L_i > lim\,d$, $lim d$ being the distance between $A_i$ and the base.

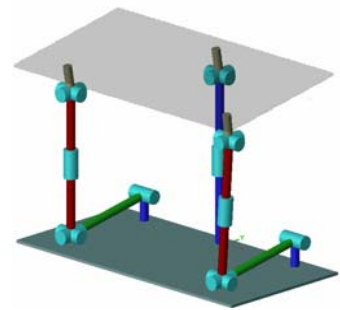

**Figure 17: The intersection between segments ($A_iB_i$) and the base**

## The algorithm used to determine the workspace

After defining all the elements necessary for the determination of the workspace, we will describe the algorithm used to accomplish this work. This algorithm consists in the following steps.

**The upper part of the workspace:**

- Initialize two matrices $W\varphi_u$ and $W_{\theta u}$, dimensions $(n_{\psi i}/2+1) n_\varphi$ where $n_{pssi}+1$ is the odd number of equidistant plans ψ with $\psi = [-180°\ 180°]$ on which the workspace will be calculated, and $n_\varphi$ is the number of points that must be calculated over each ψ. These matrices record respectively values φ and θ for each point defined in the upper part of the border of the workspace.
- Let ψ = 0, we assume $(\varphi_C\ \theta_C) = (0°\ 0°)$ is the center of the horizontal cross-section of the workspace for ψ= 0°.
- For the current ψ, build a polar coordinate system $(\varphi_C\ \theta_C)$. Starting with $n_\varphi$ equally spaced angles, and increasing the polar radius to solve the inverse kinematic problem, and test the validity of solutions by checking the constraints until a point appears where at least one constraint is broken. The last valid value (φ, θ) is saved in $W\varphi_u$ and $W_{\theta u}$ as the border of the workspace.
- Calculate the geometric center $(\varphi_C\ \theta_C)$ of the workspace cross-section to use it as a center for the next cross-section.
- Increase ψ so that $\psi = \psi + 360°/n_\psi$ while $\psi < 360°$ or the last horizontal cross-section of the workspace is a single point.

**The lower part of the workspace:**

- Initialize two matrices $W\varphi_l$ and $W_{\theta l}$, dimensions $(n_{\psi i}/2+1) n_\varphi$.
- Let $\psi = 0°$. Assume $(\varphi_C\ \theta_C) = (0°\ 0°)$ is the center of the horizontal cross-section of the workspace for ψ=0°.
- Proceed in the same manner as in the upper part to check the collision.
- Calculate the geometric center $(\varphi_C\ \theta_C)$ of the cross-section of the working space that will serve as the geometric centre for the next section.
- Take $\psi = \psi - 360°/n_\psi$.
- Repeat until ψ is smaller than -180° or the last horizontal cross-section of the workspace is a single point.

**Treatment of values to plot the workspace:**

Let $W\varphi = [W\varphi_u\ W\varphi_l]$ and $W_\theta = [W_{\theta u}\ W_{\theta l}]$, we build the workspace in a polar coordinate system as

$$x[i, j] = W_\theta[i, j] \cos(W_\varphi[i, j])$$
$$y[i, j] = W_\theta[i, j] \sin(W_\varphi[i, j])$$
$$z[i, j] = \psi_{max} - (i-1)(360°/n_\psi)$$

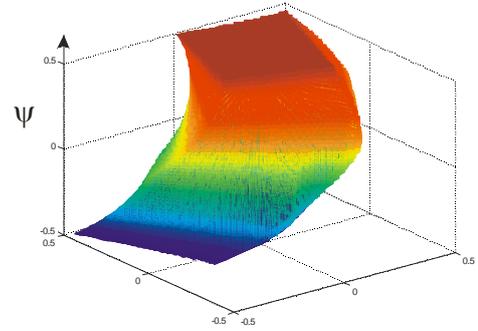

**Figure 18: Workspace without any collisions**

The cross-section of the workspace permits to check its symmetry for opposition value of ψ as is shown in Fig. 19 for $\psi = -18°$ and $\psi = 18°$. The size of the cross-section decreases if we chose $\psi < -18°$ or $\psi > 18°$.

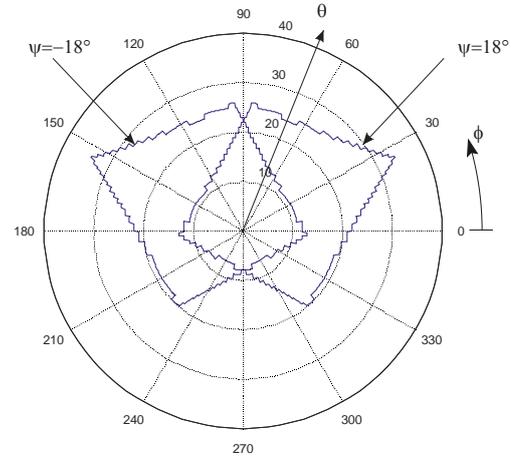

**Figure 19: Cross-section of the workspace for $\psi = -18°$ and $\psi = 18°$**

For $\psi = 0°$, we have the maximal size of the cross-section as shown in Fig. 20.

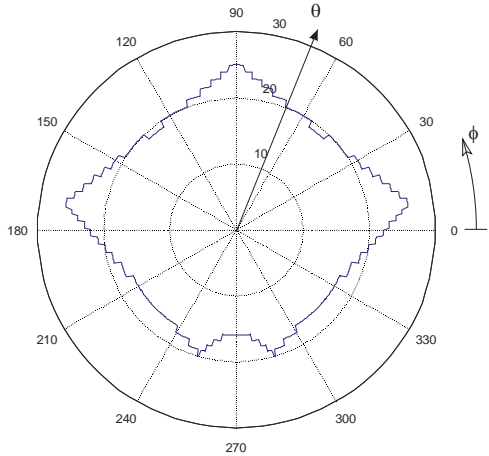

**Figure 20: Cross-section of the workspace for $\psi = 0°$**

The torsion of the mobile plat-form is $\psi = \begin{bmatrix} -18° & 18° \end{bmatrix}$ and the range values of the actuated joints are:

$$\theta_1 = \begin{bmatrix} -17° & 38° \end{bmatrix}$$
$$\theta_2 = \begin{bmatrix} -17° & 38° \end{bmatrix}$$
$$\theta_3 = \begin{bmatrix} -35° & 35° \end{bmatrix}$$

With cad software, we can build a surface passing through the border of the joint space by filtering the data obtained with the inverse kinematics problem (Fig. 21). We can easily apply an offset on this surface to define a security distance.

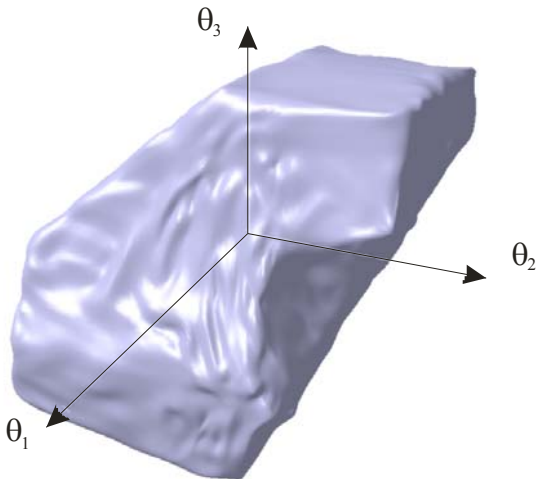

**Figure 21: The joint space modelling with a CAD software**

## CONCLUSIONS

In this paper, we have presented the kinematic equations as well as the singular configurations of a spherical wrist able to be a part of an eel-like robot. An algorithm is written to compute the workspace without any internal collisions. The Tilt-and-Torsion parameters are used to represent the workspace. Thanks to these parameters, we are able to see its symmetrical property conversely to Euler parameters.

## ACKNOWLEDGMENTS

This research was partially supported by the CNRS ("Anguille" Project) and ANR RAAMO. The author wishes to thank the technical staff of the laboratory to have realized the prototype, Michaël Canu, Gaël Branchu, Fabrice Brau and Paul Maulina as well as my master student, Gerges Fadel.